\documentclass[final]{cvpr}

\usepackage{times}
\usepackage{epsfig}
\usepackage{graphicx}
\usepackage{amsmath}
\usepackage{amssymb}
\usepackage{caption}
\usepackage{mathrsfs}
\usepackage{multirow}
\usepackage{caption}

\usepackage[pagebackref=true,breaklinks=true,colorlinks,bookmarks=false]{hyperref}

\def\cvprPaperID{6862} 
\def\confYear{CVPR 2021}
\begin{document}

\title{BiCnet-TKS: Learning Efficient Spatial-Temporal Representation for Video Person Re-Identification}
\author{Ruibing Hou$^{1,2}$,  Hong Chang$^{1,2}$, Bingpeng Ma$^{2}$, Rui Huang$^{3}$, Shiguang Shan$^{1,2,4}$\\
$^1$Key Laboratory of Intelligent Information Processing of Chinese Academy of Sciences (CAS),\\Institute of Computing Technology, CAS, Beijing, 100190, China\\
$^2$University of Chinese Academy of Sciences, Beijing, 100049, China\\
$^3$Shenzhen Institute of Artificial Intelligence and Robotics for Society, \\ The Chinese University of Hong Kong, Shenzhen, Guangdong, 518172, China\\
$^4$CAS Center for Excellence in Brain Science and Intelligence Technology, Shanghai, 200031, China\\
{\tt\small ruibing.hou@vipl.ict.ac.cn, bpma@ucas.ac.cn, ruihuang@cuhk.edu.cn, \{changhong, sgshan\}@ict.ac.cn}
}
\maketitle

\begin{abstract}
In this paper, we present an efficient spatial-temporal representation for video person re-identification (reID). Firstly, we propose a Bilateral Complementary Network (BiCnet) for spatial complementarity modeling. Specifically, BiCnet contains two branches. Detail Branch processes frames at original resolution to preserve the detailed visual clues, and Context Branch with a down-sampling strategy is employed to capture long-range contexts. On each branch, BiCnet appends multiple parallel and diverse attention modules to discover divergent body parts for consecutive frames, so as to obtain an integral characteristic of target identity. Furthermore, a Temporal Kernel Selection (TKS) block is designed to capture short-term as well as long-term temporal relations by an adaptive mode.  TKS can be inserted into BiCnet at any depth to construct BiCnet-TKS for spatial-temporal modeling. Experimental results on multiple benchmarks show that BiCnet-TKS outperforms state-of-the-arts with about $50\%$ less computations. The source code is available at \url{https://github.com/blue-blue272/BiCnet-TKS}.
\end{abstract}

\section{Introduction}
Person re-identification (reID)~\cite{PCB,mars,IANet} aims at retrieving a particular person across multiple non-overlapped cameras.  Recently, with the emergence of large video benchmarks~\cite{mars,GLTL} and the growth of computational resource, video person reID has been attracting a lot of attention. The video data contain richer spatial and temporal clues, which can be utilized to reduce visual ambiguities for more robust reID. 

\begin{figure}[t]
\centering
   \includegraphics[width=0.95\linewidth]{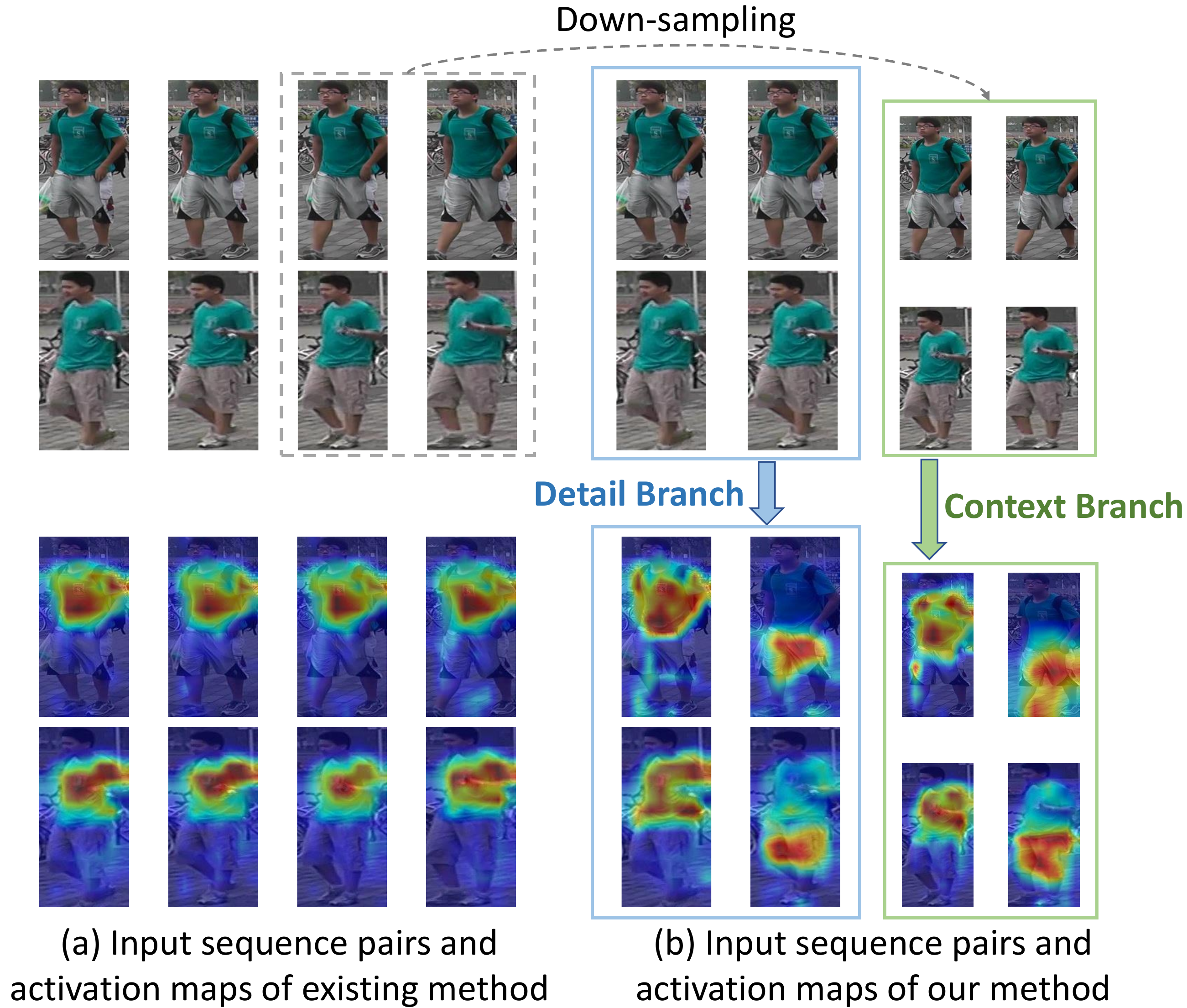}
   \captionsetup{font={small}}
   \caption{An example of class activation maps~\cite{CAM} of a pair of input video sequences of existing method~\cite{mars} and our method. }
\label{motivation}
\vspace*{-1.0em}
\end{figure}

Despite the significant progress in video reID, most existing methods do not take full advantage of the rich spatial-temporal clues in videos. For \textit{spatial clues}, most methods~\cite{RCN,QAN,VRSTC} conduct the same operation on each frame at same input resolution, resulting in highly redundant spatial features for consecutive frames. The redundant features easily focus on the same most representative local region~\cite{TCLNet}, which may be indistinguishable for the two persons with seemingly similar local body parts. For example, as shown in Fig.~\ref{motivation} (a), the green T-shirt of the sequence pair attracts the most attention, but is difficult to distinguish the two pedestrians. Therefore, it is desirable to automatically capture the diverse spatial clues across consecutive frames to form a full characteristic of each identify.

\begin{figure}[t]
\centering
   \includegraphics[width=0.9\linewidth]{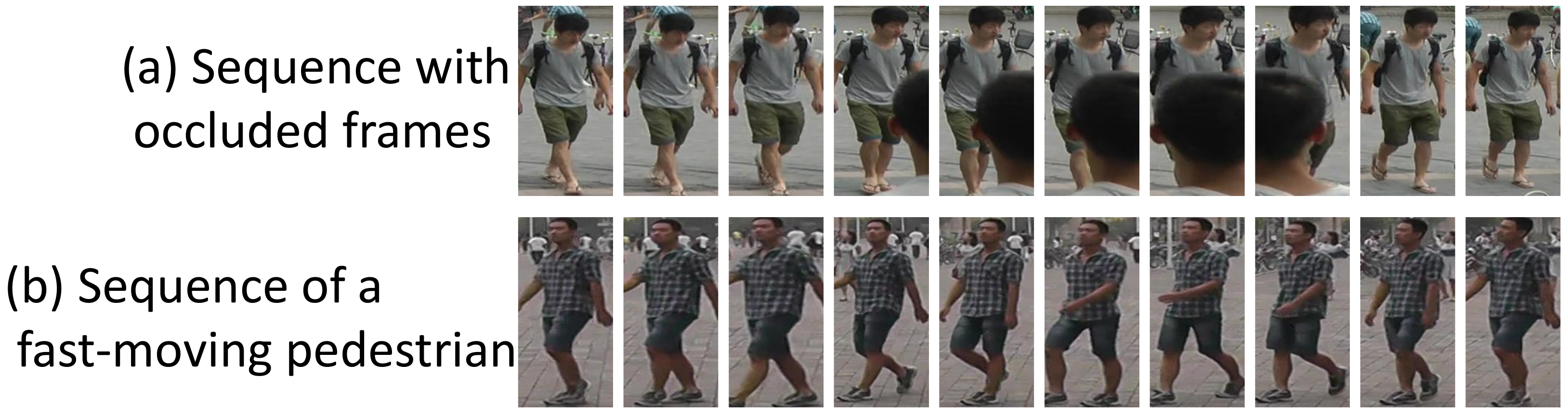}
   \captionsetup{font={small}}
   \caption{Short and long-term temporal relations have varying importance for different sequences. (a) A  sequence with partial occlusion. The long-term temporal clues are desired to alleviate occlusion.  (b) A sequence of a fast-moving pedestrian. The short-term temporal clues are desired to model detailed motion patterns.} 
\label{motivation-1}
\vspace*{-1em}
\end{figure}

For \textit{temporal clues}, most existing methods only model either short-term~\cite{RCN,yan2016person,Gu3D} or long-term temporal relations~\cite{non-local,yang2020spatial,IAUnet}. To enhance the temporal modeling ability, a few works~\cite{GLTL,M3D} attempt to jointly capture short and long-term temporal relations and fuse the two relations with equal weights. However, the two temporal relations have varying importance for different sequences. For example, as shown in Fig.~\ref{motivation-1}, for a sequence with partial occlusion, the long-term temporal relations are more important to alleviate occlusion. For a fast-moving pedestrian sequence, the short-term temporal relations play a greater role to model the detailed motion patterns. So it is necessary to \textit{adaptively} capture short and long-term temporal relations of videos.

To explicitly fulfill above goals, we present an efficient spatial-temporal representation for video reID. We first propose a \textit{Bilateral Complementary Network} (BiCnet) to extract complementary spatial features across consecutive frames. \textbf{Firstly}, BiCnet contains two scale-specific branches, \textit{Detail Branch} operating on frames at original resolution to retain spatial details, and \textit{Context Branch} processing frames at down-sampled resolution to enlarge receptive field for long-range contexts. As shown in Fig.~\ref{motivation} (b), with larger receptive field, the third-frame feature of the first sequence can capture broader visual clues of a green T-shirt with a backpack strap on it, which can help differentiate the two similar pedestrians. 
\textbf{Then} on each branch, BiCnet appends multiple parallel spatial attention modules. By enforcing the diversity of individual attention modules, the attention modules can focus on different regions for consecutive frames. As shown in Fig.~\ref{motivation} (b), with the diverse attention modules, the consecutive-frame features from same branch can focus on complementary body regions, covering the whole body of the target identity.  \textbf{Finally}, BiCnet aggregates the complementary features from the two branches to a comprehensive spatial representation.

Furthermore, we develop a \textit{Temporal Kernel Selection} (TKS) block to \textit{adaptively} model the short and long-term temporal relations. Utilizing both small kernel and large kernel along the temporal dimension can capture the short and long-term temporal relations simultaneously. So TKS is designed to contain several parallel temporal convolution paths with various kernel sizes. More importantly, TKS selects a dominant temporal scale according to the global information from the multiple paths. With the selection strategy, TKS can adaptively vary the scale of temporal modeling depending on the properties of input videos, thereby exhibiting stronger temporal representational capability. TKS is computationally lightweight and imposes a slight increase in model complexity. It can be readily inserted into BiCnet, called ``BiCnet-TKS'', to progressively learn spatial-temporal patterns. 

We evaluate our approach on multiple challenging video reID benchmarks. The evaluations show that our approach outperforms state-of-the-arts. Moreover, by down-sampling some frames to low-resolution, BiCnet-TKS greatly reduces the computations, requiring about $50\%$ less computation cost than state-of-the-arts.

\section{Related Work}

\textbf{Person ReID.}\quad 
Existing video reID methods mainly focus on exploiting rich spatial-temporal clues in videos. For spatial clues, most works~\cite{mars,jointly,QAN,AD-zhao} apply temporal average pooling or a weighting strategy to fuse frame features. For temporal clues, existing methods use optical flow~\cite{RCN,See,jointly}, recurrent neural network~\cite{RCN,yan2016person,tran2015learning,snippet}, 3D convolution~\cite{V3DP,Gu3D} or non-local block~\cite{non-local,VRSTC,IAUnet} to model the temporal relations. Recently, the works~\cite{M3D,GLTL} propose to jointly capture short and long-term temporal relations. However, these methods fuse the two temporal relations with equal weights. In contrast, our TKS adaptively selects a dominate temporal relation based on the input video, exhibiting stronger temporal modeling capability.

 The most similar work to ours BiCnet is TCLNet~\cite{TCLNet}, which also extracts complementary features for consecutive frames. BiCnet has several advantages over it. First, TCLNet only considers one spatial scale to focus on local details, while our method is built on a two-branch architecture, which can capture both detailed features as well as long-range contexts.  Second, TCLNet uses \textit{hard erasing} to drop the salient features which may deteriorate the representation capacity, while our method adopts \textit{soft attention} to flexibly determine the regions that should be attended to. Third, TCLNet uses multiple expensive CNNs to mine diverse parts. Our method uses diverse and lightweight attention modules with sharing CNNs, which is more computational efficient and parametric friendly.

\label{sec3}
\begin{figure*}[t]
\begin{center}
\captionsetup{font={small}}
\centerline{\includegraphics[width=0.99\textwidth]{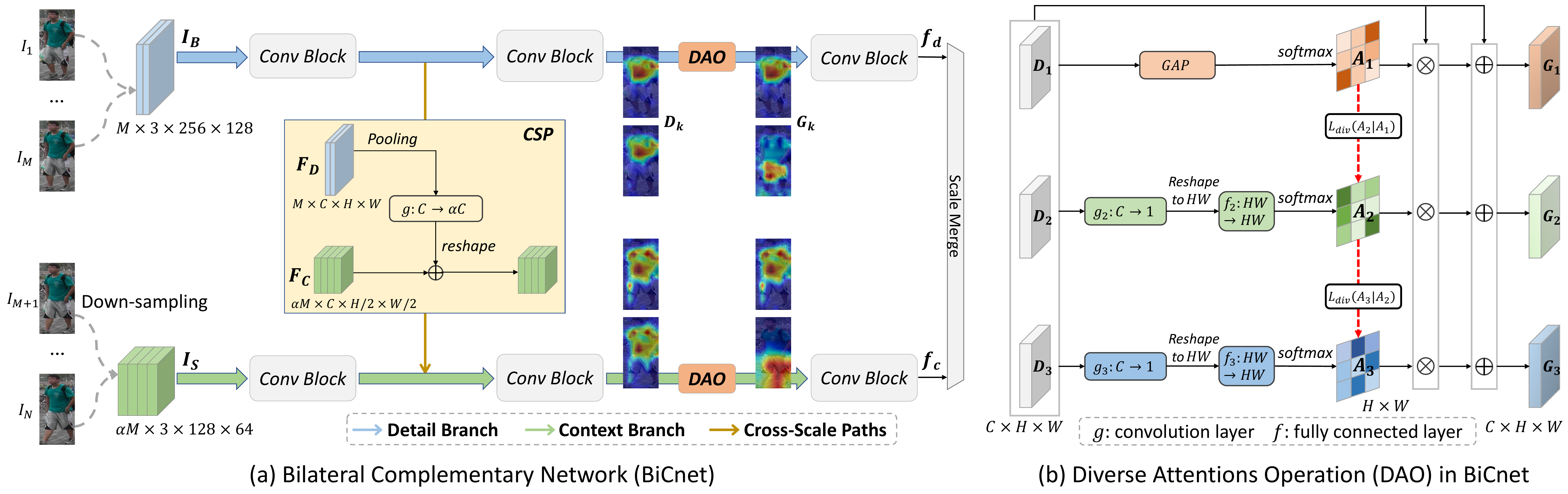}}
\vspace*{-0.5em}
\caption{(a) The overall framework of BiCnet.  BiCnet contains two branches. \textit{Detail Branch} processes  frames at original resolution to encode detailed spatial clues, and \textit{Context Branch} processes frames at half of  original resolution to provide larger receptive field for long-range contexts. The input frames of a sequence are split into different branches. \textit{Cross-Scale Paths} (CSP) fuse the two branches after each stage. \textit{Diverse Attentions Operation} (DAO) is added on each branch which enforces consecutive frames to focus on different body regions, so as to obtain an integral characteristic of each identity. (b) The structure of DAO where a three-frame case is shown, and the divergence term $L_{div}(A_3|A_1)$ is omitted for clarity. }
\label{BiNet}
\end{center}
\vspace*{-2.0em}
\end{figure*}

\textbf{Multi-branch Architecture.}\quad 
Multi-branch architecture has exhibited great success in image based vision tasks. For example, M3DNet~\cite{MSD} and HR-Nets~\cite{sun2019deep} propose the networks that contain multiple branches and each branch has it own spatial resolution, respectively for image classification and pose estimation. The works~\cite{chen2017person,liu2016multi} propose a pyramidal feature learning network  that consists of multiple scale-specific feature learning branches for image reID. However, above methods process each image at multiple resolutions, incurring additional computations. On the contrary, our approach uses an individual resolution for each frame which largely reduces the computational cost. Moreover, very few methods explore the multi-branch architecture for efficient video understanding. SlowFast Networks~\cite{feichtenhofer2019slowfast} rely on a similar two-branch structure, but each branch encodes different frame rates, while our method processes frames with different spatial resolutions.

\textbf{Attention Model.}\quad 
Attention mechanism has proven to be a potential way to enhance CNNs. SENet~\cite{hu2018squeeze} proposes an efficient channel attention module. 
CBAM~\cite{woo2018cbam} and BAM~\cite{park2018bam} further introduce spatial attention block. SKNet~\cite{li2019selective} brings the feature attention across two spatial convolutions. Recent methods~\cite{wang2020eca,dai2020attentional,zhang2020resnest,hu2018gather} further improve the channel attention block.
However existing methods are usually designed to enhance spatial representational capability. In contrast, our TKS adopts attention over different temporal kernels, which can boost the temporal representational power of video networks. Also, our BiCnet is the first work to use \textit{diverse attention modules} across \textit{consecutive frames} to enhance the video representation.

\section{Our Approach}
We aim at developing an efficient spatial-temporal representation for video reID. Our method includes two novel components, \textsl{i.e.}, BiCnet for complementary spatial representations across consecutive frames, and TKS for adaptively modeling the short and long-term temporal relations.  

\subsection{Bilateral Complementary Network}

As shown in Fig.~\ref{motivation} (a), most existing methods extract highly redundant features for consecutive frames that only highlight a local body part~\cite{TCLNet}. To this end, we design a  Bilateral Complementary Network to mine complementary visual clues from consecutive frames. As shown in Fig.~\ref{BiNet}, BiCnet is built on a two-branch architecture and adds a Diverse Attentions Operation (DAO) on each branch. The two-branch architecture is used to model complementary scales for different video sub-segments, and DAO is utilized to mine complementary body parts for consecutive frames. By adding DAO on each branch, BiCnet can obtain an integral characteristic of the target person, producing a comprehensive spatial representation.

\textbf{Two-branch Architecture.}\quad 
As shown in Fig.~\ref{BiNet} (a), BiCnet contains two CNN branches, a Detail Branch processing former several frames of given video segment at original resolution and a Context Branch operating on remaining frames at half of original resolution. By down-sampling input frames to small size,  Context Branch provides larger receptive field to encode long-range spatial contexts, which can complement the detailed features extracted by Detail Branch. Concretely, suppose a video segment $I=\{I_n\}_{n=1}^N$ contains $N$ consecutive frames and $n$ is the index of the video frame.  We firstly divide $I$ into two sub-segments, namely big frames $I_B=\{I_n\}_{n=1}^{M}$ at original resolution, and small frames $I_S=\{I_n\}_{n=M+1}^N$ at half of the original  resolution, where $M$ is a hyper-parameter that determines the ratio of the small frames to big frames $\alpha$. Then $I_{B}$ and $I_{S}$ are fed into Detail Branch ($\text{CNN}_{\text{D}}$) and Context Branch ($\text{CNN}_{\text{C}}$) separately, to obtain the corresponding feature vectors $f_{d}$ and $f_{c}$  as follows,
\begin{equation}
\begin{split}
f_{d} &= \frac{1+\alpha}{N}\sum_{k} \text{CNN}_{\text{D}}(I_{k}), I_{k} \in \{I_n\}_{n=1}^{\frac{N}{1+\alpha}} \\
f_{c} &= \frac{1+\alpha}{\alpha N} \sum_{k} \text{CNN}_{\text{C}}(I_{k}), I_{k} \in  \{I_n\}_{n=\frac{N}{1+\alpha}+1}^{N}.
\end{split}
\label{eq1}
\end{equation}
Finally, we simply average $f_{d}$ and $f_{c}$ to obtain the video feature for recognizing. 

\textbf{Cross-Scale Paths.}\quad 
Further, we add Cross-Scale Paths (CSP) that propagate the intermediate information of Detail Branch to Context Branch. CSP enables Context Branch to aware the features extracted by Detail Branch, such that Context Branch can focus on exploiting long-range visual clues less activated by the other branch.

The structure of CSP is illustrated in Fig.~\ref{BiNet} (a). Formally, let $F_{D}\in \mathbb{R}^{M\times C \times H \times W}$ and $F_{C}\in \mathbb{R}^{\alpha M\times C \times \frac{H}{2} \times \frac{W}{2}}$  be the intermediate video feature map extracted by the same stage of Detail Branch and Context Branch respectively, where $C,H$ and $W$ denote the number of channels, the height and the width of feature map of big frames respectively. $F_D$ and $F_C$ have different spatial and temporal dimensions, so CSP first performs transformation on $F_D$ to $\overline{F_D}\in \mathbb{R}^{\alpha M\times C \times \frac{H}{2} \times \frac{W}{2}}$ to match the size as:
\begin{equation}
\overline{F_D} = \mathcal{R}\left(W_c * \mathcal{P}(F_D)\right).
\end{equation}
Here $\mathcal{P}$ is the pooling operation that performs max pooling with stride 2 to match the spatial dimension, $*$ is the convolution operation, $W_c\in\mathbb{R}^{1\times 1 \times C \times \alpha C }$ is the parameter of the convolution operation, and $\mathcal{R}$ is the reshape operation reshaping the convolutional result with size $M \times \alpha C \times\frac{H}{2} \times \frac{W}{2} $ to $\alpha M \times C \times \frac{H}{2} \times \frac{W}{2}$ to math the temporal dimension. At last, $\overline{F_D}$ is fused into  $F_C$ by element-wise summation.

\textbf{Diverse Attentions Operation.} \quad
As shown in Fig.~\ref{BiNet} (a), although the big frames and small frames can provide some complementary clues (\textit{e.g.}, detailed T-shirt/additional long-distance knapsack strap feature), the frames on each branch still easily focus on around the most representational region (\textit{e.g.}, upper-clothes). To this end, we design Diverse Attentions Operation to mine complementary regions for consecutive frames. By adding DAO on each branch, BiCnet can discover abundant discriminative parts and produce an integral complementary characteristic of each identity. 

As shown in Fig.~\ref{BiNet} (b), DAO contains several parallel attention modules and uses a specific attention module for each frame. By encouraging diversity among the generated attention maps, the attention modules can attend to complementary parts, so as to acquire diverse discriminative features for the consecutive frames. 

In particular, DAO takes $F_D$ (or $F_C$) as input, and uses a specific attention module for each frame feature map $(F_D)_k\in\mathbb{R}^{C\times H\times W}$. We take $F_D$ as an example, 
and denote $(F_D)_k$ as $D_k$ for simplicity.  Firstly, as pointed out by~\cite{choe2019attention}, the intensity of each pixel in high-level feature map is proportional to the discriminative power. So we compress $D_1$ by channel-wise average pooling to locate the region activated by $D_1$, producing a self-attention map  $A_1\in\mathbb{R}^{H\times W}$:
\begin{equation}
A_k = \textit{softmax}\left(\frac{1}{C} \sum_{c=1}^{C} (D_k)_c\right),\quad k=1,
\end{equation}
Then we introduce parallel  attention modules to learn to mine different and non-activated regions. Specifically, given $D_k$ ($k$$>$1), the corresponding attention module first takes a convolutional layer to compress the channel dimension and reshapes the result to $\mathbb{R}^{HW}$. After that a fully-connected layer is applied to embed the global spatial contexts. Finally, the result is reshaped to $\mathbb{R}^{H\times W}$ followed by a softmax layer to produce corresponding attention map $A_k\in\mathbb{R}^{H\times W}$ ($k$$>$1).

In order to guide different  attention modules to activate diverse regions, the corresponding spatial attention maps should be different. To achieve this, a  \textit{divergence regularization term} is introduced to measure the diversity of two attention maps $A_k$ and $A_l$, which is defined as:
\begin{equation}
L_{div}(A_k|A_l) = 1 - \textit{sim}(A_k, A_l),
\label{eq5}
\end{equation}
where $\textit{sim}(A_k, A_l)$ computes the similarity of $A_k$ and $A_l$. Any distance measure is applicable, and we use the dot-product similarity~\cite{non-local} since dot-product is more implementation-friendly in modern deep learning platforms. Then the divergence loss is calculated as:
\begin{equation}
L = \frac{-1}{M-1} \sum_{k=2}^{M} \left(\frac{1}{k-1} \sum_{l=1}^{k-1} L_{div}\left(A_k|A_{l}\right)\right).
\end{equation}
$L$ is used to guide the optimization of parallel attention modules. When any two attention modules focus on similar person region, the generated attention maps would have a low diversity value, producing a high loss value $L$. So optimizing with $L$ can drive the different attention modules to focus on different person regions. Next, we encode the diverse attention information into input feature maps by a residual operation. 

At last, the updated feature maps are fed into the subsequent convolutional layers to generate feature vectors embedded with complementary visual clues.

\subsection{Temporal Kernel Selection Block}
\label{sec4}
Following~\cite{qiu2017learning,xie2018rethinking}, we factor the video network to treat spatial clues and temporal relations separately. With the efficient BiCnet to fully mine the spatial clues, we build a \textit{Temporal Kernel Selection} block to jointly model the short-term and long-term temporal relations. Since the temporal relations with different scales have varying importance for different sequences (as illustrated in Fig.~\ref{motivation-1}), TKS combines the multi-scale temporal relations in a dynamic way, \textit{i.e.}, different weights are assigned to different temporal scales according to input sequences.

In particular, TKS takes a sequence of consecutive-frame feature maps $F=\{F_t\}_{t=1}^T$ as input, where $F_t$ is the feature map of the $t^{th}$ frame, and conducts a triple of operations, \textit{Partition}, \textit{Select} and \textit{Excite} on $F$.  

\textbf{\textit{Partition} Operation.} Due to imperfect person detection algorithm, the adjacent frames of a video are not well aligned, which might make the temporal convolution ineffective on video reID~\cite{Gu3D}. Following~\cite{PCB}, we use the partition strategy to alleviate the spatial misalignment issue. Specifically, given video feature map $\{F_t\}_{t=1}^T$, we divide each frame feature map into $h \times w$ spatial regions uniformly, and perform average pooling on each divided region to construct a region-level video feature map $X\in\mathbb{R}^{T\times C\times h \times w}$.

\begin{figure}[t]
\centering
   \includegraphics[width=0.95\linewidth]{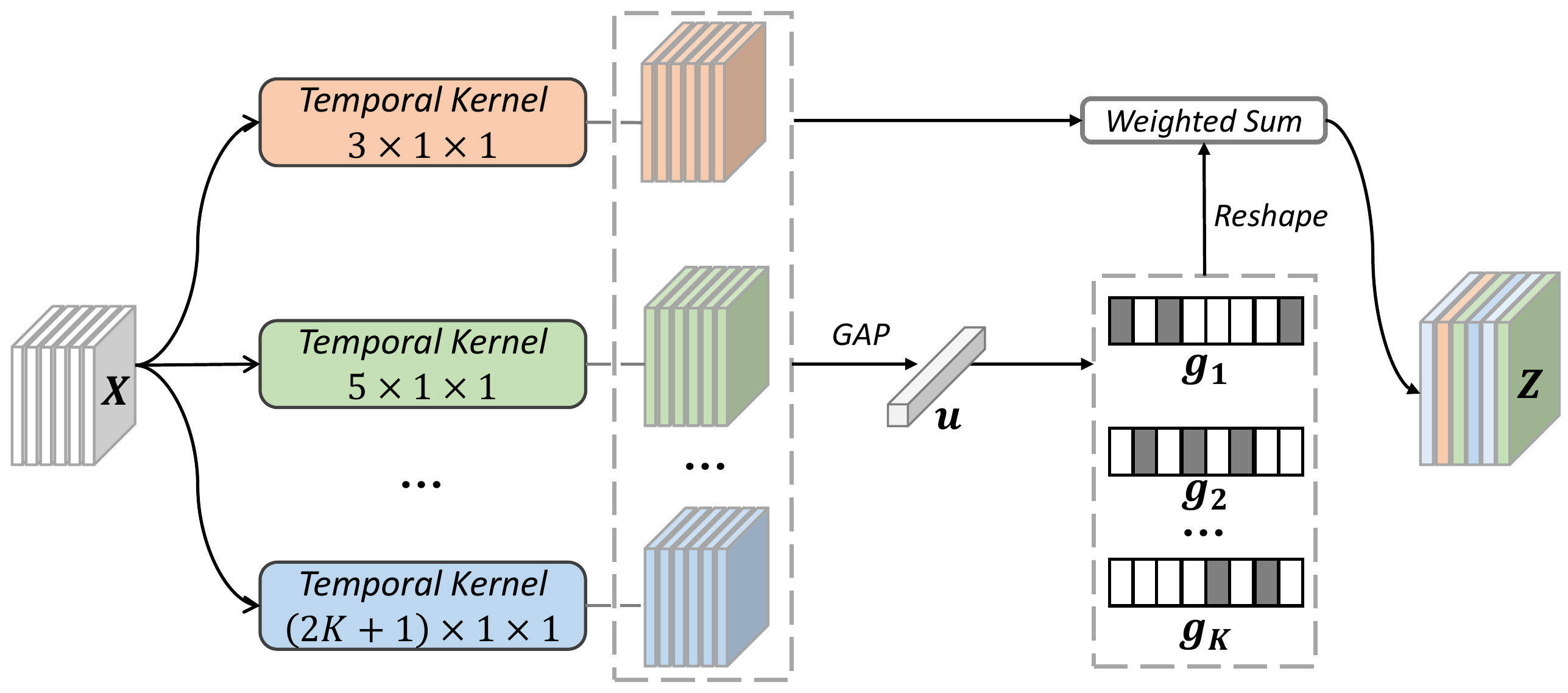}
   \captionsetup{font={small}}
   \caption{The architecture of Temporal Kernel Selection block.}
\label{TKS}
\vspace*{-0.5em}
\end{figure}  

\textbf{\textit{Select} Operation.} As shown in Figure.~\ref{TKS}, given $X$, we conduct $K$ parallel paths $\{\mathcal{F}^{\left(i\right)}: X \rightarrow Y^{\left(i\right)} \in\mathbb{R}^{T\times C\times h \times w}\}_{i=1}^K $, where $\mathcal{F}^{\left(i\right)}$ is 1D temporal convolution~\cite{qiu2017learning} with kernel size $2i+1$. For further efficiency, the temporal convolution with a $\left(2i+1\right) \times 1\times 1$ kernel is replaced with dilated convolution with a $3\times 1\times 1$  kernel and dilation size $i$. 
The basic idea of \textit{select} operation is to use global information from all temporal paths to determine the assigned weights to each path. In particular, we first fuse the outputs of all paths by element-wise summation, then perform global average pooling to obtain a global feature $u\in\mathbb{R}^{C\times 1}$:
\begin{equation}
u = \textit{GAP}_{T,h,w}(\sum_{i=1}^K Y^{\left(i\right)}),
\label{eq4}
\end{equation}
where  $\textit{GAP}_{T,h,w}$ denotes global average pooling along the temporal and spatial dimension. After that the channel selection weights $\{g_i\in\mathbb{R}^{C\times 1}\}_{i=1}^K$ are obtained according to the global embedding $u$,
\begin{equation}
g_i = \frac{\exp\left( W_iu \right)}{\sum_{j=1}^K \exp \left(W_ju\right)} \quad i\in\{1,\dots,K\},
\label{eq7}
\end{equation}where $W_i\in\mathbb{R}^{C\times C}$ is the transformed parameters to generate $g_i$ for  $Y^{\left( i \right)}$. 
The aggregated feature map $Z\in\mathbb{R}^{T\times C\times h\times w}$ is then obtained through the selection weights on various temporal kernels,
\begin{equation}
Z=\sum_{i=1}^K \mathcal{R}(g_i) \odot Y^{\left(i\right)},
\label{eq8}
\end{equation}where $\mathcal{R}$ is the reshape operation reshaping $g_i\in \mathbb{R}^{C\times 1}$ to
$\mathbb{R}^{1\times C \times 1 \times 1}$ to be compatible with the size of $Y^{\left(i\right)}$. 

It is worth pointing out that, in contrast to using scale-wise weight to provide coarse fusion, we choose to use channel-wise weights (Eq.~\ref{eq7}) for fusing. This design results in more fine-grained fusion that tunes each feature channel. In addition, the weights are dynamically computed conditioned on input videos. This is crucial for reID where different sequences may have different dominate temporal scales.

\textbf{\textit{Excite} Operation.} The \textit{excite} operation modulates the input feature map by conditioning on $Z$ with a residual scheme. The final feature map $E=\{E_t\}_{t=1}^T$ is obtained as: $E_t=\mathcal{U}\left(Z_t\right) + F_t$. Here $\mathcal{U}$ is the nearest neighbor upsampler that performs upsampling on $Z_t$ to match the spatial resolution of $F_t$. TKS block maintains the input size, thus can be inserted at any depth of BiCnet to extract efficient spatial-temporal feature.

\subsection{Overall Architecture}
\label{overall}

Our idea of BiCnet is generic, and it can be instantiated with different backbones~\cite{szegedy2015going,szegedy2016inception,residual}. Following recent works~\cite{Gu3D,GLTL,Co-segmentation}, we use ResNet-50~\cite{residual} pretrained on ImageNet~\cite{krizhevsky2017imagenet} with last down-sampling operation removed as the backbone. The branches of BiCnet are built on ResNet-50 that consists of four consecutive stages, \textit{i.e.}, \textit{stage}$_1$$\sim$\textit{stage}$_4$. Diverse Attentions Operation is added after \textit{stage}$_3$ since the high-level feature maps contain more semantic information. TKS block can be inserted into BiCnet to any stage to construct BiCnet-TKS for spatial-temporal modeling. 

\textbf{Structure and Weight Sharing between Branches.} 
An immediate problem of multi-branch architecture~\cite{chen2017person} is that it introduces several times parameters and incurs a higher risk of overfitting. 
So we use the same structure and share the parameters for the two branches of BiCnet. It reduces the number of parameters and makes BiCnet need no extra parameters over single-branch reID network.

\begin{table*}[t]
\captionsetup{font={small}}
\caption{Comparison with state-of-the-arts on MARS, DukeMTMC-VideoReID and  LS-VID datasets. The methods are separated into three groups, mainly for spatial (\textbf{S}), temporal (\textbf{T}) and spatial-temporal (\textbf{ST}) modeling.}
\small
\centering
\begin{tabular}{c | l | c c | c c| c c}
\hline
\multicolumn{2}{c|}{\multirow{2}*{Methods}} & \multicolumn{2}{|c|}{MARS}  &\multicolumn{2}{|c|}{Duke-Video} &\multicolumn{2}{|c}{LS-VID}\\   
\cline{3-8}
\multicolumn{2}{c|}{ } &mAP &top-1 &mAP & top-1& mAP &top-1 \\
\hline
\multirow{2}*{\textbf{S}}
&COSAM*~\cite{Co-segmentation} &79.9 &84.9 &94.1 & 95.4 &-  &- \\
&MGRAFA~\cite{zhang2020multi} &85.9 &88.8 &- &- &-&-\\
\hline
\multirow{4}*{\textbf{T}}
&Two-stream~\cite{simonyan2014two} &- &- &- &- & 32.1 &48.2 \\
&STMP~\cite{liu2019spatial} &72.7 &84.4 &- &- & 39.1 &56.8 \\
&M3D~\cite{M3D}  &74.1 &84.4 &- &- &40.1 &57.7 \\ 
&GLTP~\cite{GLTL}  & 78.5 & 87.0 &93.7 &96.3& 44.3 &63.1  \\
\hline
\multirow{9}*{\textbf{ST}} 
&DRSA~\cite{diversity} &65.8 &82.3 &- &- &37.8 &55.8 \\
&VRSTC~\cite{VRSTC} & 82.3 & 88.5 &93.5  &95.0 &- &-   \\
&I3D~\cite{carreira2017quo} &83.0 &88.6 &- &- &33.9 &51.0\\
&P3D~\cite{qiu2017learning} &83.2 &88.9 &- &- &35.0 &53.4\\
&STGCN~\cite{yang2020spatial} &83.7 &89.9 &95.7 &\textbf{97.3} &- &- \\
&IAUnet~\cite{IAUnet} &85.0 &\textbf{90.2}&96.1&96.9 &-&-\\
&TCLNet~\cite{TCLNet} &85.1 &89.8 &\textbf{96.2} &96.9 &70.3 &81.5 \\
&AP3D~\cite{Gu3D} & 85.1 & 90.1 &95.6  &96.3 &73.2 &84.5  \\
&MGH~\cite{yan2020learning} &85.8 &90.0 &- &- &- &- \\
\hline
\multirow{1}*{\textbf{ST}}
&BiCnet-TKS  &\textbf{86.0} &\textbf{90.2} &96.1 &96.3 &\textbf{75.1} &\textbf{84.6} \\
\hline
\end{tabular}
\label{tab-s}
\end{table*}

\textbf{Computation Cost Analysis.} 
To illustrate the computation cost of BiCnet-TKS, we consider a common video reID Baseline~\cite{mars} that uses ResNet-50 to extract feature for each frame at original resolution.  Assume that the FLOPs for Baseline to extract one-frame feature is $p$, Baseline requires $Np$ FLOPs to process a video with $N$ frames. BiCnet-TKS splits the video frames to big frames at original resolution and small frames at half of original resolution by a ratio $1:\alpha$ (Eq.~\ref{eq1}). So BiCnet-TKS requires about $\frac{N}{1+\alpha}p + (\frac{\alpha N}{1 + \alpha})\frac{p}{4}$ FLOPs\footnote{The computations of CSP, DAO and TKS are negligible compared to the feature extraction of ResNet-50.}, corresponding to about $\frac{3}{4} - \frac{3}{4\alpha+4}$ relative decrease over Baseline. 

We can see that the the computation cost decreases as $\alpha$ increases. However, when $\alpha$ is too large, the small frames would dominate the network optimization, causing a severe performance drop. We experimentally observe that setting $\alpha$ to $3$ offers the best trade-off between computation cost and accuracy. In this case, BiCnet-TKS only requires $\sim$$44\%$ computation costs over Baseline, which is more efficient to extract the spatial-temporal feature. 

\section{Experiment}
\subsection{Dataset and Settings}
\textbf{Datasets.}
We evaluate the proposed method on multiple video reID datasets, \textsl{i.e.}, MARS~\cite{mars}, DukeMTMC-VideoReID~\cite{dukevideo} and LS-VID~\cite{GLTL}.

\textbf{Evaluation Metric.} We adopt mean Average Precision (mAP)~\cite{map} and  Cumulative Matching Characteristics (CMC)~\cite{cmc}  as evaluation metrics.

\textbf{Implementation Details.} During training, for each video sequence, we randomly sample $8$ frames with a stride of four frames to form a video segment. Each batch contains $16$ persons, each person with $4$ video segments. We resize the split big frames to $256\times 128$ and small frames to $128\times 64$.  The horizontal flip and random erasing~\cite{zhong2017random} are adopted for data augmentation. As for the optimizer, Adam~\cite{adam} with weight decay $0.0005$ is adopted to update the parameters. We train the model for $150$ epochs in total. The learning rate is initialized to $3.5\times 10^{-4}$ with a decay factor $0.1$ at every $40$ epochs.  In BiCnet, the ratio of small frames to big frames is set to $3$. 
In TKS, the number of temporal kernels is set to $2$, and the divided regions is  $4\times2$. 

During testing, for each video sequence, we first split it into several 8-frame video segments. Then we extract the feature for each video segment by BiCnet-TKS and the final video feature is the averaged representation of all segments. After feature extraction, the cosine distances between the query and gallery features are computed for retrieval.

\subsection{Comparison with State-of-the-arts}
In Tab.~\ref{tab-s}, we compare our method with state-of-the-arts on MARS and DukeMTMC-VideoReID and LS-VID datasets. 
Our method achieves the best performance. It is noted that: \textbf{(1)} The spatial-based methods~\cite{Co-segmentation,Coherence,zhang2020multi} process each frame by same operation and resolution, so they do not fully consider the spatial redundancy between frames. On the contrary, our BiCnet ensures different frames to focus on divergent regions to form an integral person representation and achieves better performance.
\textbf{(2)} Our method outperforms TCLNet~\cite{TCLNet}, with an improvement up to $4.8\%$ mAP on LS-VID dataset. The significant improvements can be attributed to the use of two-branch architecture and flexible soft attention modules. \textbf{(3)} The temporal-based methods~\cite{snippet,V3DP,Gu3D} lack the ability of modeling both short and long-term temporal relations. Our method outperforms these methods with an $1\%$ mAP improvement on MARS. \textbf{(4)}. The methods~\cite{GLTL,M3D,yan2020learning} aggregate the multi-scale temporal relations with equal weights. Our method achieves better performance by an adaptive selection mechanism. \textbf{(5)}. All existing methods add computations over Baseline. In contrast, our method greatly reduces the computation cost by processing some frames at low-resolution. Overall, our method outperforms state-of-the-arts with about $50\%$ computation budgets.

\subsection{Ablation Study}
In this section, we respectively investigate the effectiveness of BiCnet and TKS block by conducting a series of ablation studies on MARS dataset. 

\subsubsection{The components of BiCnet.}
To validate the effectiveness of BiCnet, we introduce a baseline that adopts ResNet-50 with temporal average pooling to generate the video feature. The baseline processes all frames at the same resolution and is trained with cross entropy and triplet loss. We consider two baseline models, \textit{i.e.}, Base-B processing frames at original resolution ($256\times 128$), and Base-S processing frames at half of original resolution ($128\times 64$). The comparisons are shown in Tab.~\ref{tab1}.

\begin{table}[t]
\captionsetup{font={small}}
\caption{Component Analysis of BiCnet-TKS on MARS. We also report the number of average floating-point operations (GFLOPs) for one frame, and the parameter number (Param.) of the networks.}
\centering
\small
\begin{tabular}{l |c c |c  c}
\hline 
\multirow{2}*{Models}  & \multicolumn{4}{c}{MARS} \\   
\cline{2-5}
&GFLOPs. &Param.  &mAP &top-1\\
\hline
Base-S ($128\times 64$)  &1.02 &23.5M  &80.7 &87.4\\
Base-B ($256\times 128$)  &4.08 &23.5M  &85.2 &89.1\\
\hline
Two-branch (TB) &1.81 &23.5M  &84.3 &89.6 \\
TB+CSP &1.89 &27.6M &\textbf{85.0}  &89.6 \\
\hline
TB+CSP+AO (wo $L_1$)&1.89 &27.6M &85.2 &89.3\\
TB+CSP+DAO (BiCnet) &1.89 &27.6M &\textbf{85.6}  &\textbf{89.8}\\
\hline
BiCnet-TK (fix-fusion) &1.91 &29.1M &85.5 &89.6\\
BiCnet-TKS  &1.99 &29.2M &\textbf{86.0} &\textbf{90.2}\\
\hline
\end{tabular}
\label{tab1}
\end{table}

\textbf{The influence of branch number.} \quad
BiCnet is built on a two-branch architecture. It is easy to extend to multiple branches case which splits the video frames into multiple groups and uses an individual resolution for each group. In this part, we conduct an uniform split for fair comparison. The results are shown in Tab.~\ref{tab2}. From Tab.~\ref{tab2}, we have following observations: (1) Training ResNet-50 on frames at  $128\times 64$ resolution still offers reasonable accuracy, while saving $75\%$ computations (measured by floating point operations). (2) Too small input resolution  ($64\times 32$) causes severe performance degradation, with a drop up to  $21\%$ mAP. We argue that too small input size leads to serious loss of spatial details, which is difficult to distinguish pedestrians with small inter-class variations.
(3) The three-branch architecture performs worse than two-branch structure. It is likely that the branch with $64\times 32$ input resolution would disturb the optimization of network parameters. So we use a two-branch architecture, which can achieve comparable performance to Base-B with less computations.

\textbf{Two-branch architecture \textit{w.r.t} split ratio.} \quad
We then investigate the influence of the split ratio $\alpha$ (in Eq.~\ref{eq1}), \textit{i.e}, the ratio of  small frames ($128\times 64$) to big frames ($256\times 128$), to the two-branch architecture (TB). The results are shown in Tab.~\ref{tab3}. We can observe that with  $\alpha$ increases, TB greatly reduces the average computations of processing one frame. But the mAP of TB decreases as $\alpha$ increases. We argue that it is due to the lack of interaction between the two branches. In particular, the two branches of TB independently extract features, so it is difficult for one branch to learn to capture the clues ignored by the other branch. Moreover, the feature discriminative power of low-resolution frames is lower than that of high-resolution frames. So directly using low-resolution frames inevitably weakens the discrimination of final features. In addition, we observe that $\alpha$$=$3 only brings slight drop compared to $\alpha$$=$2. Considering computational complexity, we set $\alpha$ to 3 in this work. 
\begin{table}[t]
\captionsetup{font={small}}
\caption{Results of Single-branch/Multi-branch Architecture with a single resolution/different combinations of multiple resolutions inputs. Height denotes the input resolution is Height$\times$(Height/2)}
\centering
\small
\begin{tabular}{c | c | c  | c c |c  c}
\hline 
  \multicolumn{3}{c|}{Height}  & \multicolumn{4}{c}{MARS} \\   
\hline
 \textbf{256} &128&64 & GFLOPs. & Param.  &mAP &top-1\\
\hline
$\checkmark$  & & &4.08 & 23.5M &\textbf{85.2} &89.1 \\
& $\checkmark$ & &1.02 &23.5M & 80.7&87.4 \\
&  & $\checkmark$ &0.25 &23.5M &64.1 &77.4 \\
\hline
$\checkmark$ & $\checkmark$ & &2.55 &23.5M &84.8 &\textbf{89.4} \\
$\checkmark$ & $\checkmark$  & $\checkmark$ &1.76 &23.5M &79.1 &86.1 \\
\hline
\end{tabular}
\label{tab2}
\end{table}
\begin{table}[t]
\captionsetup{font={small}}
\caption{Results of Two-branch architecture (TB) with different $\alpha$ (the ratio of small frames to big frames).}
\centering
\small
\begin{tabular}{l |c c |c  c}
\hline 
\multirow{2}*{$\alpha$}  & \multicolumn{4}{c}{MARS} \\   
\cline{2-5}
&GFLOPs. &Param.  &mAP &top-1\\
\hline
 0 (Base-B) &4.08 &23.5M  &\textbf{85.2} &89.1\\
 \hline
 1  &2.57 &23.5M  &84.8 &89.4\\
 2  &2.07 &23.5M  &84.4 &\textbf{89.7}\\
 3  &1.81 &23.5M  &84.3 &89.6\\
 4  &1.67 &23.5M  &83.8 &89.5\\
\hline
 $+\infty$ (Base-S) &1.02 &23.5M &80.7 &87.4\\
\hline
\end{tabular}
\label{tab3}
\end{table}

\textbf{Effectiveness of Cross-Scale Paths.} \quad
We evaluate the effect of CSP by adding it after each stage of above two-branch architecture. As shown in Tab.~\ref{tab1}, compared with TB, employing CSP brings $0.7\%$ mAP gains with small computational overhead. We argue that with the information propagation from Detail Branch to Context Branch, Context Branch can enhance its representational power. In addition, the two branches can learn to work collaboratively to mine complementary clues, \textit{i.e.}, Detail Branch extracts the detailed feature of local body parts, and Context Branch focuses more on the long-distance contexts, to further enhance the feature representation. 

\textbf{Effectiveness of Diverse Attentions Operation.}\quad
Finally, we investigate the individual effect of the attention modules and divergence constraint on DAO. The results are presented in Tab.~\ref{tab1}. The difference between TB+CSP+AO and TB+CSP+DAO is that TB+CSP+AO appends parallel attention modules without $L_1$ to guide optimization. As shown in Tab.~\ref{tab1}, TB+CSP+AO achieves negligible gains over TB+CSP, which indicates that the visual features captured by different attention modules are almost the same. TB+CSP+DAO achieves $0.6\%$ mAP improvement over TB+CSP, which validates the capability of the proposed divergence regularization term. We argue that the divergence loss enforces different attention modules to focus on complementary person regions and form an integral characteristic of target identity. The integral characteristic is more conductive to distinguish different identities with similar local parts. 

\subsubsection{The components of TKS block.}
\begin{table}[t]
\captionsetup{font={small}}
\caption{Results of BiCnet-TKS with different combinations of multiple temporal kernels in TKS. }
\centering
\small
\begin{tabular}{c | c | c |c c |c  c}
\hline 
\multicolumn{3}{c|}{kernel size }  & \multicolumn{4}{c}{MARS} \\   
\hline
 K3& K5& K7 &GFLOPs. &Param.  &mAP &top-1\\
\hline
$\checkmark$ & & &1.94 &28.3M &85.1 &89.9 \\
& $\checkmark$  & &1.94 &28.3M &85.3 &90.1 \\
 & & $\checkmark$ &1.94 &28.3M &85.5 &89.8 \\
\hline
$\checkmark$ & $\checkmark$ & &1.99 &29.2M &\textbf{86.0} &\textbf{90.2} \\
 $\checkmark$ & &$\checkmark$  &1.99 &29.2M &85.7 &90.0 \\
 & $\checkmark$ & $\checkmark$  &1.99 &29.2M &85.6 &90.1 \\
\hline
 $\checkmark$ & $\checkmark$ & $\checkmark$ &2.04 &30.0M &85.8 &90.2 \\
 \hline
\end{tabular}
\label{tab4}
\end{table}

\textbf{Effectiveness of TKS.} \quad
We first assess the effectiveness of TKS block by adding it after \textit{stage}$_2$ of BiCnet in Tab.~\ref{tab1}. TKS brings $0.4\%$ mAP and top-1 accuracy gains over BiCnet with an extremely small increase in computational complexity. We argue that TKS is complementary to BiCnet, \textit{i.e.}, TKS provides the temporal features that cannot be extracted by BiCnet. Furthermore,
in order to verify the effect of the adaptively selection mechanism in TKS, we introduce a \textit{Temporal Kernel} (TK) block which simply averages the results with the multi-scale kernels ($Z=\frac{1}{K}\sum_{i=1}^K Y^{(i)}$ in Eq.~\ref{eq8}). As shown in Tab.~\ref{tab1}, TK brings no gain over BiCnet, which indicates that the improvement of BiCnet is attributed to the adaptive selection among the multi-scale kernels.

\textbf{TKS \textit{w.r.t} the number of temporal kernels ($K$).} \quad
Next, we investigate the influence of combination of different kernels. We consider three different kernels, called ``K3'' (standard $3\times1\times1$ 3D convolutional kernel), 	``K5'' ($3\times1\times1$ convolution with dilation 2 to approximate $5\times1\times1$ kernel size), and ``K7'' ($3\times1\times1$ convolution with dilation 3 to approximate $7\times1\times1$ kernel size). The results are shown in Tab.~\ref{tab4}. We can observe that: (1) When using two temporal kernels with different sizes, in general the accuracy increases. The mAP and top-1 accuracy in the second block of the table ($K=2$) are generally higher than those in the first block ($K=1$), indicating the effectiveness of modeling both short and long-term temporal relations. (2) Using more temporal kernels ($K=3$) does not bring performance gain, showing two temporal kernels are enough to capture the temporal clues of video. 

\textbf{Efficient positions to place TKS.} \quad
\begin{table}[t]
\captionsetup{font={small}}
\caption{Results of BiCnet-TKS when placing TSK blocks on different stages.}
\centering
\small
\begin{tabular}{l |c c |c  c}
\hline 
\multirow{2}*{Stage}  & \multicolumn{4}{c}{MARS} \\   
\cline{2-5}
&GFLOPs. &Param.  &mAP &top-1\\
\hline
\textit{stage}$_{1}$ &1.99 &28.0M  &85.3 &90.1\\
\textit{stage}$_{2}$ &1.99 &29.2M  &\textbf{86.0} &90.2\\
\textit{stage}$_{3}$ &1.99 &34.1M  &85.7 &\textbf{90.4}\\
\textit{stage}$_{4}$ & 2.29&53.5M  &85.4 &90.0\\
\hline
\textit{stage}$_{23}$ &2.09 &35.7M  &85.8 &90.3\\
\hline
\end{tabular}
\label{tab5}
\end{table}
Tab~\ref{tab5} compares the results of placing a TKS block to different stages of BiCnet. It can be seen that the improvements by placing one TKS block in \textit{stage}$_2$ and \textit{stage}$_3$ are similar. However, placing TKS block in \textit{stage}$_1$ and \textit{stage}$_4$ leads to performance degradation. It is likely that the low-level features in \textit{stage}$_1$ are insufficient to provide precise semantic information, thus TKS can not model temporal relations between body parts very well. And since BiCnet learns to focus on different regions for consecutive frames on \textit{stage}$_3$, the frame features on \textit{stage}$_4$ lack of coherent temporal relations, so TKS is not capable to extract an effective temporal feature on \textit{stage}$_4$. 
We also observe that adding more TKS blocks does not bring gain, indicating that a TKS block is usually enough for temporal modeling.

\textbf{Time Overhead}.  \quad
The running times are positively correlated with computation cost of models. In Tab.~\ref{tab1}, Base-B takes 11ms to extract feature for a 8-frames sequence. While BiCnet-TKS only takes 6ms, corresponding to a $45.4\%$ relative decrease over Base-B (both timings are performed on one NVIDIA 2080Ti GPU).

\section{Conclusions}
In this work, we present a computation-friendly spatial-temporal representation for video reID. Firstly, we introduce Bilateral Complementary Network. BiCnet contains two branches, Detail Branch preserving the spatial detail clues from original resolution, and Context Branch utilizing down-sampling operation to enlarge receptive field for longer-range contexts modeling. On each branch, BiCnet appends parallel and diverse attention modules to mine divergent regions for consecutive frames.
Furthermore, we propose Temporal Kernel Selection block to adaptively capture temporal relations of videos. Extensive experiments demonstrate the superiority of our method over state-of-the-arts with about $50\%$ less computations.

\noindent\textbf{Acknowledgement} 
This work is partially supported by Natural Science Foundation of China (NSFC): 61876171 and 61976203, and the Open Project Fund from Shenzhen Institute of Artificial Intelligence and Robotics for Society, under Grant No. AC01202005015 and 2019-INT006.

{\small
\bibliographystyle{ieee_fullname}
\bibliography{egbib}
}

\end{document}